\documentclass[10pt,twocolumn,letterpaper]{article}

\usepackage{iccv}
\usepackage{times}
\usepackage{epsfig}
\usepackage{graphicx}
\usepackage{amsmath}
\usepackage{amssymb}
\usepackage{algorithm}
\usepackage{algorithmic}
\usepackage{subfigure}
\usepackage{multirow}
\usepackage{graphicx}
\usepackage{newfloat}
\usepackage{listings}
\usepackage{booktabs}
\usepackage{bm}
\usepackage{amsmath}
\usepackage{amsfonts}
\usepackage{amsthm}
\usepackage{booktabs}
\usepackage{graphicx}
\usepackage{bm}
\usepackage{amsmath}
\usepackage{adjustbox}
\usepackage[final]{pdfpages}
\usepackage[breaklinks=true,bookmarks=false]{hyperref}

\iccvfinalcopy % *** Uncomment this line for the final submission

\def\iccvPaperID{****} % *** Enter the ICCV Paper ID here

\ificcvfinal\fi

\begin{document}

%%%%%%%%% TITLE
\title{ECM-OPCC: Efficient Context Model for Octree-based Point Cloud Compression}

\author{Yiqi Jin\textsuperscript{\rm 1,2}\thanks{This work is done when Yiqi Jin and Yuhuan Lin are interns at AIR, Tsinghua University},
Ziyu Zhu\textsuperscript{\rm 3},
Tongda Xu\textsuperscript{\rm 1},
Yuhuan Lin\textsuperscript{\rm 1,2}, 
Yan Wang\textsuperscript{\rm 1}\thanks{Yan Wang is the corresponding author.} \\
\textsuperscript{\rm 1}Institute for AI Industry Research (AIR), Tsinghua University \\\textsuperscript{\rm 2}Department of Automation, Tsinghua University \\\textsuperscript{\rm 3}Department of Computer Science, Tsinghua University}

\maketitle
\ificcvfinal\thispagestyle{empty}\fi

%%%%%%%%% ABSTRACT
\begin{abstract}
   Recently, deep learning methods have shown promising results in point cloud compression. For octree-based point cloud compression, previous works show that the information of ancestor nodes and sibling nodes are equally important for predicting current node. However, those works either adopt insufficient context or bring intolerable decoding complexity (e.g. $>$ 900s). To address this problem, we propose a sufficient yet efficient context model and design an efficient deep learning codec for point clouds. Specifically, we first propose a segment-constrained multi-group coding strategy to exploit the autoregressive context while maintaining decoding efficiency. Then, we propose a dual transformer architecture to utilize the dependency of current node on its ancestors and siblings. We also propose a random-masking pre-train method to enhance our model. Experimental results show that our approach achieves state-of-the-art performance for both lossy and lossless point cloud compression. Moreover, our multi-group coding strategy saves 98\% decoding time compared with previous octree-based SOTA compression method \cite{fu2022octattention}.
\end{abstract}

%%%%%%%%% BODY TEXT
\section{Introduction}
\label{sec:intro}
% With the rapid development of 3D computer vision, point cloud has growing applications in many fields (\textit{e.g.} virtual reality\cite{surveydl}, autonomous driving\cite{xu2022opv2v} and robotics \cite{zhang2022pointclip}). On the other hand, effective point cloud storage and transmission has also became an emerging challenge as amount of point cloud data grows. For example, in \textit{vehicle-to-vehicle} (V2V) communications \cite{wang2020v2vnet}, real-time communication of point cloud data across vehicle is required. For map applications, millions of points generated by LiDAR \cite{wiesmann2021deep} has to be stored. To wrap up, effective point cloud compression is required by the industry applications.

% The \textit{moving picture experts group} (MPEG) has developed several standards like VPCC and GPCC to compress point cloud by hand-crafted modules \cite{surveydl,pointcloudsurvey,overview}. Both 3D octree structure and 2D projection are exploited in these standards to adapt to different scenarios.

Recently, Deep Point Cloud Compression (DPCC) has outperformed traditional hand-crafted codec \cite{fu2022octattention}. Inspired by learned image compression and traditional point cloud codec, deep learning community invents tree-based, voxel-based, projection-based and point-based compression algorithms: Tree-based approach represents point cloud by octree \cite{octsq} and aggregates ancestor and sibling information to predict current node occupancy; Voxel-based approach quantizes point cloud into voxel grid \cite{que2021voxelcontext} and uses 3D convolution as learned transform; Projection-based approach \cite{tu2019point} projects 3D point cloud to 2D image and uses image compression techniques to compress it; And point-based approach \cite{you2021patch} utilizes modules from point cloud understanding \cite{liang2022transpcc} to transform point cloud and compress transformed points and features. Among above-mentioned methods, octree-based approach is commonly used for large scale sparse point cloud \cite{pointcloudsurvey}.

\begin{figure}[t]
    \centering
    \includegraphics[width=\columnwidth]{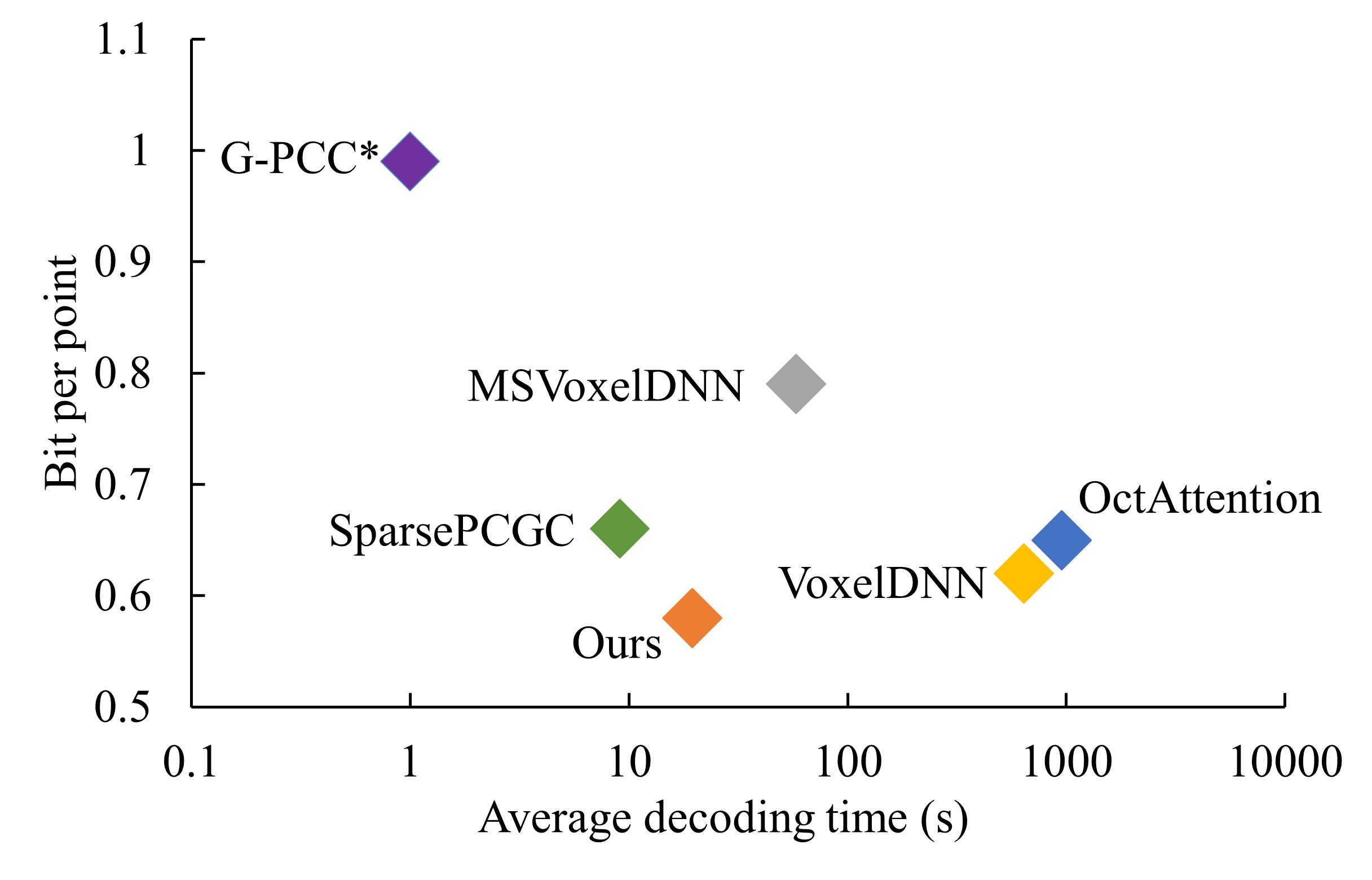}
    \caption{The bpp-decoding time of different methods. Our method achieves SOTA bpp while shows faster or comparable decoding time compared with other deep learning methods. *G-PCC is traditional non-deep learning method.}
    \label{bpptime}
\end{figure}

A major issue of octree-based methods is the decoding complexity due to the autoregressive context model. For example, in \cite{fu2022octattention}, both ancestor and sibling information are introduced as autoregressive context models, which greatly increases the decoding complexity. Inspired by parallel autoregressive density estimation \cite{pmlr-v70-reed17a}, we propose to solve this problem by multi-group coding strategy. Specifically, we factorize the fully autoregressive context of \cite{fu2022octattention} into layer-wise autoregressive context, and slice the nodes within the same layers into parallel-decodable context segment. Within each context segment, we deploy group-wise autoregressive context.
We propose a dual transformer architecture to process branches containing ancestor and sibling information. Besides, we introduce random masking pre-training strategy to boost the performance. Experimental results show  our method outperforms previous deep point cloud compression methods regarding compression ratio and obtains very competitive decoding speed (Fig.~\ref{bpptime}). 

Our main contributions are as follows:
\begin{itemize}
    \item We propose a new segment-constrained multi-group coding strategy that enables parallel decoding of nodes inside each group, which greatly accelerates decoding process while maintaining compression performance.
    \item We propose a novel dual transformer architecture with level-parallel and group-parallel branch to better extract context information from ancestors and siblings. 
    \item We introduce randomly masking input occupancy code as an efficient pre-training strategy to boost performance of our context model.
    \item Our proposed model achieves SOTA bitrate for lossless compression and R-D (Rate-Distortion) performance for lossy compression. Moreover, it saves 98\% decoding time compared with previous work \cite{fu2022octattention}.
\end{itemize}

%------------------------------------------------------------------------
\section{Related Work}

\subsection{Learned Point Cloud Compression}
Voxel-based approaches ~\cite{convtransform, improveddeep} quantize point cloud to voxel grids and use 3D auto-encoders to compress them. VoxelDNN ~\cite{nguyen2021learning,nguyen2021lossless, nguyen2021multiscale} introduces hybrid compression by first building octree to store structural information of point cloud and then compressing local voxelized point cloud by 3D convolution. Another line of research ~\cite{wang2021multiscale, wang2021sparse} also uses hybrid scheme  but downsamples point cloud by learned-transform first and then utilizes tree-based approach to compress point cloud.

For sparse point cloud, projection based approaches ~\cite{tu2019point,sun2019novel,feng2020real,zhao2022real} first project point cloud to image then use image compression methods to compress them. \cite{riddle, wang2022point} directly compress the range images of LiDAR data to leverage the lidar scanning pattern. This kind of method handles large sparse point cloud well and evolves as image compression methods improve. The above mentioned point cloud compression methods can compress point cloud in lossy and lossless way but do not preserve density in different regions of a point cloud. To tackle this problem, point-based methods ~\cite{you2021patch,he2022density,huang20193d,liang2022transpcc} utilize network designed for raw point cloud segmentation as transform and encode transformed points and features by arithmetic coding. 

\subsection{Octree-based Point Cloud Compression}
% In general, there are four types of DPCC frameworks: tree-based, voxel-based, projection-based and point-based approach. 
Octree-based approaches \cite{octsq,muscle} represent point cloud by octree data structure and convert point cloud compression to lossless tree structure compression.
%3D scenes constructed by point cloud is quantized and then partitioned recursively into 8 octants for a single node. Each octant has an 8-bit occupancy symbol, which should be transmitted for the convenience of decoding.%
Specifically, OctSqueeze \cite{octsq} uses deep recurrent model to aggregate ancestral information inside octree to predict occupancy of children node. MuSCLE \cite{muscle} improves this work by introducing temporal information and traverses octree from both top and bottom for better compression ratio. %Both these two works use recurrent neural network (RNN) to aggregate features but ignore spatial relation for each octant. 
VoxelContext-Net  \cite{que2021voxelcontext} extracts voxel representation and utilizes 3D convolution based deep entropy model to compress nodes in octree. To further increase receptive field, OctAttention \cite{fu2022octattention} gathers information from sibling and ancestor nodes in a fully autoregressive manner, and introduces attention mechanism for point cloud compression. Aside from the performance enhancement by this fully autoregressive model, the decoding complexity is greatly increased. \cite{pccSiblingContextAndSurfacePriors} fit quadratic surfaces with a voxel-based geometry-aware module to provide geometric priors in entropy encoding.

\subsection{Efficient Context Model}
As we stated above, the autoregressive context in octree-based DPCC greatly increases the decoding complexity ($>900$s per point cloud), which hinders the practical deployment of such methods. On the other hand, advancements in parallelizable autoregressive model in density estimation \cite{pmlr-v70-reed17a} and language models \cite{gu2017non,ghazvininejad2019mask} provide valuable insight for efficient context modeling in deep data compression. In the field of deep image compression, several works try to reduce decoding time by smartly design parallelizable autoregressive context model. Checkerboard context model \cite{he2021checkerboard} decodes half latent codes first and uses them as context for the other half. This idea is later adopted in deep video compression for fast decoding \cite{li2022hybrid}. ELIC \cite{he2022elic} further improves channel context model by unevenly grouping. However, efficient parallelizable autoregressive context for DPCC is under-explored, which requires considering the uniqueness of point cloud data.

\section{Methodology}
\subsection{Background: Octree-based Compression}

\begin{figure}[t]
    \centering
    \includegraphics[width=\columnwidth]{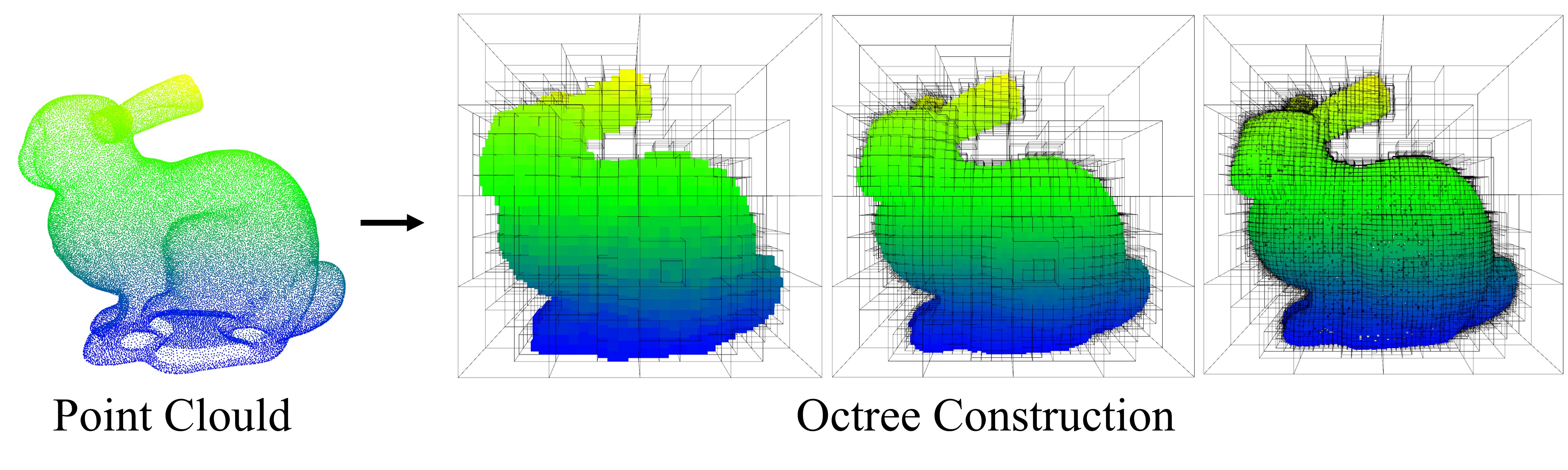}
    \caption{Point cloud BunnyMesh converted to octree with max depth of 5, 6, 7. The accuracy increases with the depth of octree.} %Octree-based context provides larger receptive field than voxel-based methods. }
    \label{bunny}
\end{figure}

\begin{figure*}[t]
    \centering
    \includegraphics[width=\textwidth]{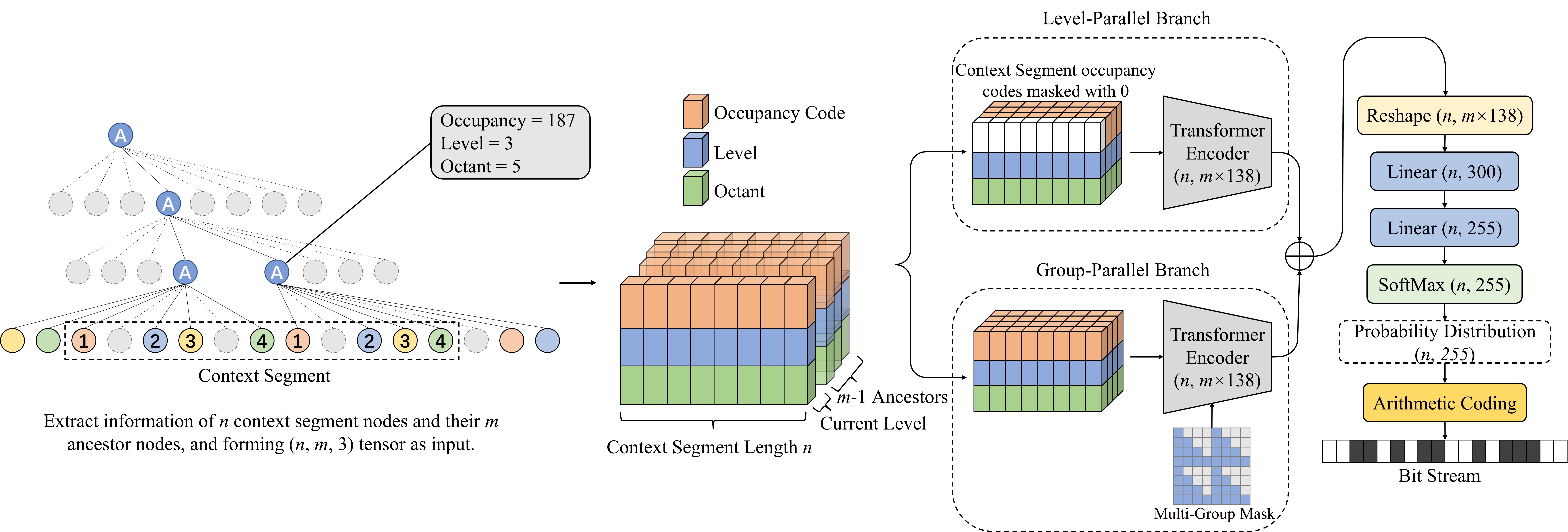}
    \caption{The overall architecture of ECM-OPCC.}
    \label{complete_procedure}
\end{figure*}

Octree is a common data structure to describe 3D space \cite{octsq}. Each node of octree represents a cube in space. And each node can have eight cubic children that subdivide this node into eight trigrams. The occupancy code of a node is a binary flag to indicate spatial occupancy. An octree can be represented as the occupancy code sequence of every node. Formally, denote the occupancy code sequence of the whole octree as $X$, we have
\begin{align}
    X=\{\bm{x}^1,...,\bm{x}^l\}
    \label{eq:ot}
\end{align}
where $l$ is the total level of octree, and $\bm{x}^i$ is the occupancy code of level $i$. The sequence of occupancy codes $\{\bm{x}^1,...,\bm{x}^l\}$ could be used to reconstruct the octree losslessly. The essence of octree-based DPCC is to losslessly compress $X$. To achieve this, we use a model parameterized by $\theta$ to construct a parametric probability distribution $P_{\theta}(X)$ to approximate the true density $P(X)$. The smaller the KL divergence between the true density $P(X)$ and the predicted probability $P_{\theta}(X)$ is, the smaller the actual compressed bitrate $\mathbb{E}_{P(x)}[-\log P_{\theta}(X)]$ we can achieve. Thus, our optimization for bitrate is Eq.~\ref{eq:mle}.

\begin{align}
    \theta^* \leftarrow \arg \min_{\theta} \mathbb{E}_{P(X)}[-\log P_{\theta}(X)]
    \label{eq:mle}
\end{align}

The distortion of octree-based DPCC only comes from the conversion between octree and point cloud. During the construction of an octree with depth $l$, a point $c_i$ in raw point cloud $C$ is quantized into the nearest octree cube. While reconstructing octree back to point cloud $\hat{C}$, the coordinate of $c_i$ is dequantized as the center of cube $\hat{c_i}$, which brings quantization error $e$. More specifically, $e$ is bounded as Eq. \ref{pcerror}, where $w$ is the max length of leaf cubes, which is determined by $l$. A deeper octree indicates finer subdivision up to arbitrary precision (See Fig. \ref{bunny}) to achieve lossless compression.
\begin{equation}
    e=\max _{i}\left\|\hat{c}_{i}-c_{i}\right\|_{\infty} \leq \frac{w}{2}
    \label{pcerror}
\end{equation}

Na\"ively constructing the density of $P_{\theta}(X)$ as a joint distribution is obviously intractable due to the extremely high dimension of $X$. Therefore, the key to octree compression is to construct a tractable factorization to $P_{\theta}(X)$ by utilizing the domain knowledge of dependency. One possible factorization is the layer-wise autoregressive method:
\begin{align}
    P_{\theta}(X)=P_{\theta}(\bm{x}^1)\prod_{i=2}^{l}P_{\theta}(\bm{x}^i|\bm{x}^{<i})
\label{eq:lw_old}
\end{align}

Without breaking this layer-wise factorization, we can further use node level and octant as auxiliary information to aid the prediction of occupancy codes. These auxiliary information in current level $i$ can be obtained from the occupancy code of previous level $i-1$. Formally, denote the auxiliary information of level $i$ as $\bm{y}^{i}$, we have $\bm{y}^i=f(\bm{x}^{i-1})$. And the auxiliary information can be used to aid the layer-wise autoregressive model as Eq.~\ref{eq:lw}:
\begin{align}
    P_{\theta}(X)=P_{\theta}(\bm{x}^1)\prod_{i=2}^{l}P_{\theta}(\bm{x}^i|\bm{x}^{<i},\bm{y}^{\le i})
\label{eq:lw}
\end{align}

Eq.~\ref{eq:lw}  is a more general formulation for OctSqueeze \cite{octsq}, where the conditional relationship on previous layers is limited within the ancestor set of current node and constrained by ancestor order, instead of the whole $\bm{x}^{<i}$. Note that these constrains do not affect the asymptotic serial decoding complexity, thus OctSqueeze \cite{octsq} is a layer-wise factorization method in nature.
This layer-wise factorization is very efficient as only information from previous layer is used as context, node within each layer can be decoded in parallel.

To fully exploit the context of ancestors and siblings, one can also adopt node-wise autoregression inside each layer $\bm{x}^i$ to obtain a probabilistic model as Eq.~\ref{eq:fam}, where $N^i$ is the length of node in layer $i$, $\bm{x}^i_j$ is the $j^{th}$ node in $i^{th}$ layer. And we call this type of model as fully autoregressive model. 

\begin{align}
    P_{\theta}(\bm{x}^i|\bm{x}^{< i},\bm{y}^{\le i})&=\notag\\ P_{\theta}(\bm{x}^i_1&|\bm{x}^{< i},\bm{y}^{\le i})\prod_{j=2}^{N^i} P_{\theta}(\bm{x}^i_j|\bm{x}^i_{<j},\bm{x}^{<i},\bm{y}^{\le i}) \label{eq:fam}
\end{align}

In fact, the factorization described by Eq.~\ref{eq:fam} corresponds to a more general formulation for OctAttention \cite{fu2022octattention}, where the conditional relationship is additionally constrained by context window and ancestor order. Note that these constrains do not affect the asymptotic serial decoding complexity, thus OctAttention \cite{fu2022octattention} is a fully autoregressive method in nature. OctAttention achieves SOTA compression performance due to its thorough capture of context information from both ancestors and siblings. However, %as decoding $\bm{x}_j^i$ requires all previous layer $\bm{x}^{<i}$ and previous node of same layer $\bm{x}^i_{<j}$, 
the decoding of such model is fully sequential. In other words, the decoding time is proportional to  $\Theta(\sum_i^l N^i)$, which can be extremely slow ($>900s$ per point cloud).
\begin{figure}[h]
    \centering
    \includegraphics[width=\columnwidth]{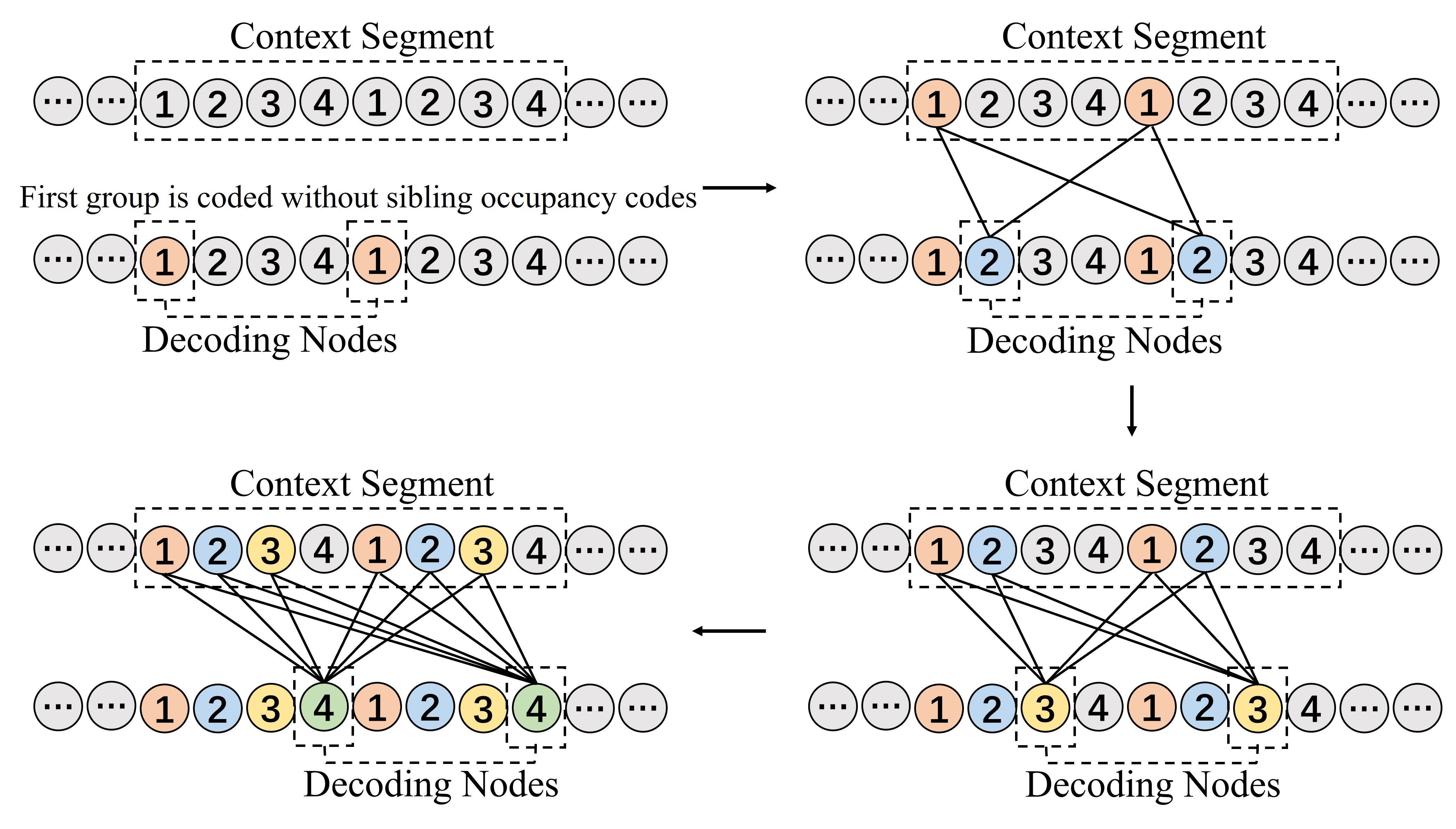}
    \caption{The decoding procedure of 4-group context segment. Nodes within same group (marked with same ID) is decoded in parallel using nodes from previous groups as context.}
    \label{groupstrategy}
\end{figure}

\subsection{Multi-Group Coding Strategy\label{sec:codingstrategy}}

As stated in Sec.~\ref{sec:intro}, we propose a multi-group coding strategy to support large-scale context while greatly improve coding efficiency over fully autoregressive model \cite{fu2022octattention}. Specifically, we compress layer $i$ of octree autoregressively, with occupancy code $\bm{x}^{<i}$, level and octant $\bm{y}^{\le i}$ from previous layers as context. Inside each layer, we divide the nodes into grouped context segment. Then we autoregressively encode each group of nodes, with previous groups inside the context segment as context (Fig.~\ref{groupstrategy}). Note that our context segment is different from the context window in OctAttention~\cite{fu2022octattention}. Our context segment divides the node in each layer into several parallel decoding segment, while context window in OctAttention is only used to reduce the range of attention calculation and there is no parallelism among different windows.

The previous layer context (ancestor) and same layer context (siblings) are modeled by our dual transformer architecture (Sec.~\ref{secdual}) to estimate the likelihood for entropy coding. In this way, the decoding process between each context segment and within each group is parallelizable, which greatly reduces decoding complexity. On the other hand, sufficient autoregressive modeling is maintained to promise compression performance. 

Formally, denote the $j^{th}$ segment of nodes in layer $i$ as $\bm{x}^i_{w_j}=\{\bm{x}^i_{w_j^1},...,\bm{x}^{i}_{w_{j}^{g}}\}$, where $g$ is the number of group, $n$ is the segment size, $h^i$ is the number of segment in $i^{th}$ layer. Then $\bm{x}^{i}_{w_j^k}$ is the $k^{th}$ group nodes of $j^{th}$ segment in $i^{th}$ layer, and we have the following factorization of layer-wise density:

\begin{align}
    P_{\theta}(\bm{x}^i|\bm{x}^{< i},\bm{y}^{\le i})=\prod_{j=1}^{h^i}P_{\theta}(\bm{x}^i_{w_j}|\bm{x}^{< i},\bm{y}^{\le i}) 
    \label{eq:wf}
\end{align}
\begin{align}
    P_{\theta}(\bm{x}^i_{w_j}|\bm{x}^{< i},\bm{y}^{\le i})&=\notag\\ P_{\theta}(\bm{x}^i_{w_j^1}&|\bm{x}^{<i},\bm{y}^{\le i})\prod_{k=2}^{g}P_{\theta}(\bm{x}^i_{w_j^k}|\bm{x}^i_{w_j^{<k}},\bm{x}^{<i},\bm{y}^{\le i})
    \label{eq:gf}
\end{align}

And taking Eq.~\ref{eq:wf} and Eq.~\ref{eq:gf} back to the layer-wise autoregressive Eq.~\ref{eq:lw}, we obtain the full likelihood model $P_{\theta}(X)$. To understand our design choice, we will discuss two questions:
\begin{itemize}
    \item How multi-group coding affects decoding time?
    \item How multi-group coding affects the bitrate?
\end{itemize}

For the first question, let's assume that our hardware is fully parallelizable (e.g. infinitely number of kernels running at the same time). For fully autoregressive model like OctAttention~\cite{fu2022octattention}, the decoding complexity is $\Theta(\sum_i^l N^i)=\Theta(|X|)$, where $|X|$ is the number of nodes in the octree. For our multi-group coding strategy, we have autoregressive dependency between layers. But for each context segment inside each layer, the decoding is parallelizable. Moreover, although the decoding between groups inside context segment is sequential, for nodes inside each group, the decoding is also parallelizable. Thus, the decoding complexity is greatly reduced to $\Theta(l \cdot g)$. And empirically, as shown in Tab.~\ref{tab:q1}, we also find our decoding time is greatly reduced over OctAttention \cite{fu2022octattention}.

For the second question, we can derive the theoretical bitrate as $\mathbb{E}_{P(X)}[-\log P(X)]+D_{KL}[P(X)||\hat{P}_{\theta}(X)]$ (See details in Appendix). And by applying the conclusion in \cite{pmlr-v162-huang22k}, we have the theoretical lowerbound on the bitrate as Tab.~\ref{tab:q2}. Although the theoretical bitrate lowerbound of our method is higher than fully autoregressive method, in practice we can achieve better bitrate than OctAttention~\cite{fu2022octattention} by smartly design the model (See Sec. \ref{secdual}) and optimization (See Sec. \ref{secpretrain}).

\begin{table}[htb]
\centering
\begin{tabular}{@{}lll@{}}
\toprule
                  & Asymptotic & Practical \\ \midrule
fully autoregressive & $\Theta(\sum_{i=1}^{l}N^i)$     & $948$s \\
Ours              & $\Theta(lg)$     &  $19.5$s \\ \bottomrule
~\\
\end{tabular}
\caption{Asymptotic and practical decoding complexity between fully autoregressive model and our proposed multi-group coding strategy assuming fully parallelizable hardware.}
\label{tab:q1}
\end{table}

\begin{table}[htb]
\centering
\begin{tabular}{@{}lll@{}}
\toprule
                  & Theoretical Lowerbound & Practical \\ \midrule
fully autoregressive & $\mathbb{E}[-\log P(X)]$     & $0.65$ bpp \\
\\
\multirow{2}{*}{Ours} & $\mathbb{E}[-\log P(X)]$     & \multirow{2}{*}{$0.58$ bpp} \\
                      & $+ \underset{ij}{\sum} \mathbb{H}(\bm{x}^i_{w_j}|\bm{x}^{<i})- \mathbb{H}(X)$ &                   \\ \bottomrule
~\\                    
\end{tabular}
\caption{Theoretical bitrate lowerbound and practical bitrate of fully autoregressive model and our proposed multi-group coding strategy.}
\label{tab:q2}
\end{table}

% Since we decode the octree by level, the complete information of ancestor nodes and the level and octant of sibling nodes could be used by all window nodes. The occupancy codes of sibling nodes are significant in predicting, we will lose a lot of accuracy without this information. Therefore, in order to balance the amount of information and coding efficiency, our approach will decode each context window by the order of groups. The occupancy codes decoded first can be used as the information of latter groups, as shown in Fig. \ref{groupstrategy}. 
% To further illustrate, the predict probability distribution of occupancy code of node $x_i$ in group k of an n-length and m-groups context window: 
% \begin{equation}
% {\hat{P}}_k(x_i)=F(x_i|a_1,a_2,...,a_n,s_1,s_2,...,s_n,\bigcup_{t=1}^{k-1}V_k)
% \label{context_eq}
% \end{equation}
% Where $a_i$ is the embedding of information of ancestor nodes of $x_i$, $s_i$ is the embedding of $x_i$, while the occupancy code is padded with zero. $V_k$ is the occupancy code decoded in previous groups. Our method only requires to call neural network n times for a n-group context window, and uses sufficient information of both ancestor nodes and sibling nodes. 
\begin{figure}[b]
    \centering
    \includegraphics[width=\columnwidth]{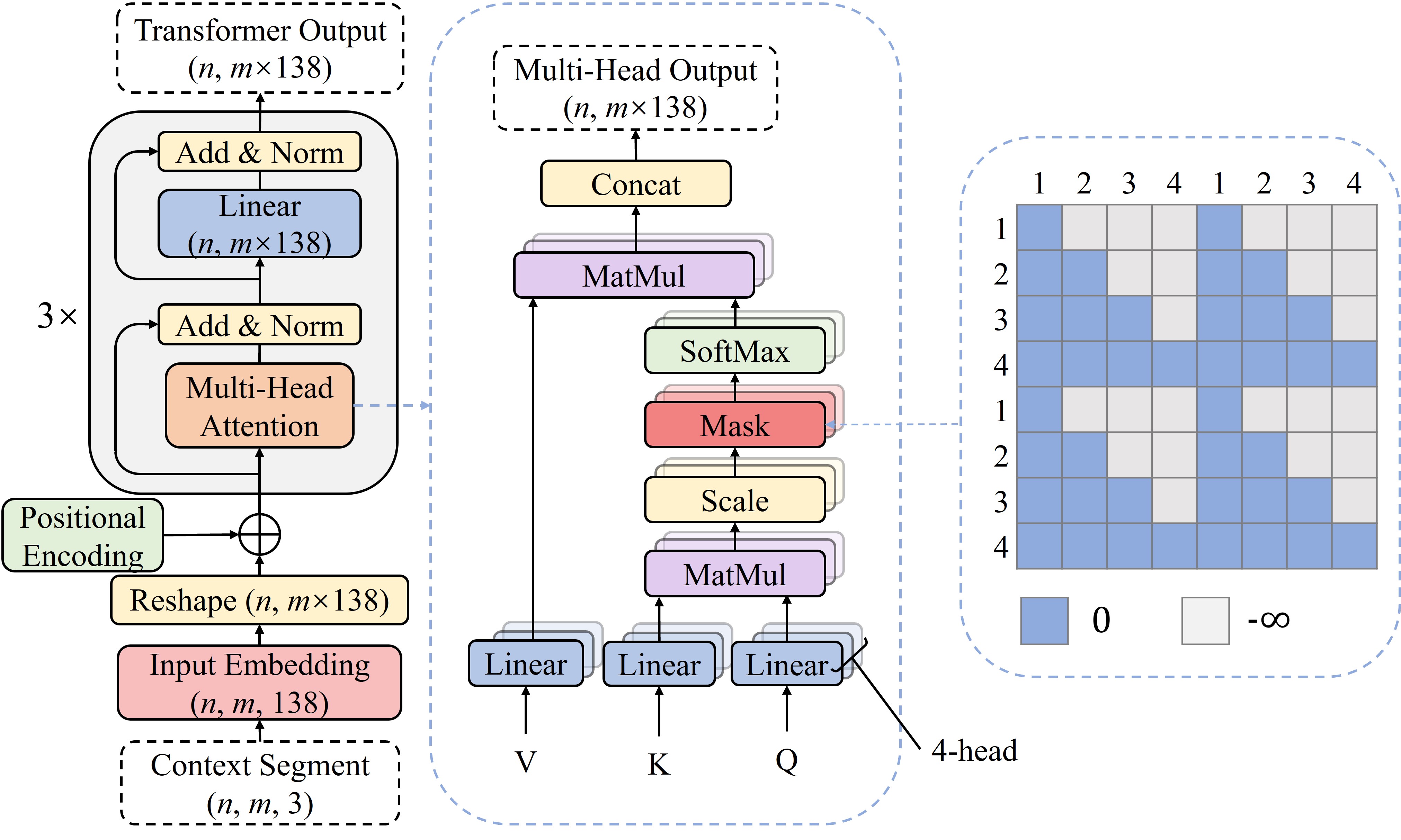}
    \caption{Transformer structure and multi-group mask matrix used in group-parallel branch. The network structure follows OctAttention \cite{fu2022octattention}. In this figure, the length of context segment $n=8$ and the group count $g=4$. The blue blocks is 0 and gray blocks is $-\infty$. }
    \label{transformerandmask}
\end{figure}

\subsection{Dual Transformer Structure\label{secdual}}
We design a novel dual branch transformer structure to support our coding strategy. Despite that our approach is fully autoregressive in layer level in nature, in practice we trace back to at most $m-1$ ancestors to save calculation (which does not affect the asymptotic complexity discussed above), where $m$ is a hyper-parameter. For each context segment $\bm{x}^i_{w_j}$ in layer $i$, we trace back $m-1$ ancestor nodes for each context segment node (pad zero if not enough), to get $n\times m$ of sibling and ancestor nodes in total. We use occupancy code $\bm{x}$, level and octant $\bm{y}$ as context to construct a $n\times m\times 3$ tensor as the input of transformer.

The level-parallel branch is designed to fully use all available information under the requirement that nodes in each level can be encoded and decoded in parallel, corresponding to Eq.~\ref{eq:lw}. As shown in Fig.~\ref{complete_procedure}, the occupancy codes of current encoding/decoding level are masked with 0 to prevent information leakage (i.e., only already decoded information can be used for current node). In this branch, the information carried by occupancy code only comes from previous levels, which are ancestor nodes.

The group-parallel branch is designed to exploit context information carried by occupancy code of sibling nodes. As shown in Fig.~\ref{transformerandmask}, we use a multi-group mask matrix to modulate the attention layer, which ensures that only already decoded occupancy code from siblings can be used as context for current node. As explained in Fig.~\ref{groupstrategy}, nodes in different context segment or nodes in the same group can be encoded or decoded in parallel, corresponding to Eq.~\ref{eq:wf} and Eq.~\ref{eq:gf}.

%By combing these two branches, we fully implement the multi-group coding strategy described in Sec.~\ref{sec:codingstrategy}. 
For encoding, we just need one forward pass of the two branches to compress all the nodes. For decoding, to decompress the node of each level, we need to run level-parallel branch once, and run group-parallel branch $g$ times, so the total sequential-calculation complexity is $\Theta(l \cdot g)$, as given in Tab.~\ref{tab:q1}.

\subsection{Random Masking Pretrain \label{secpretrain}}
\begin{figure}[t]
    \centering
    \includegraphics[width=\columnwidth]{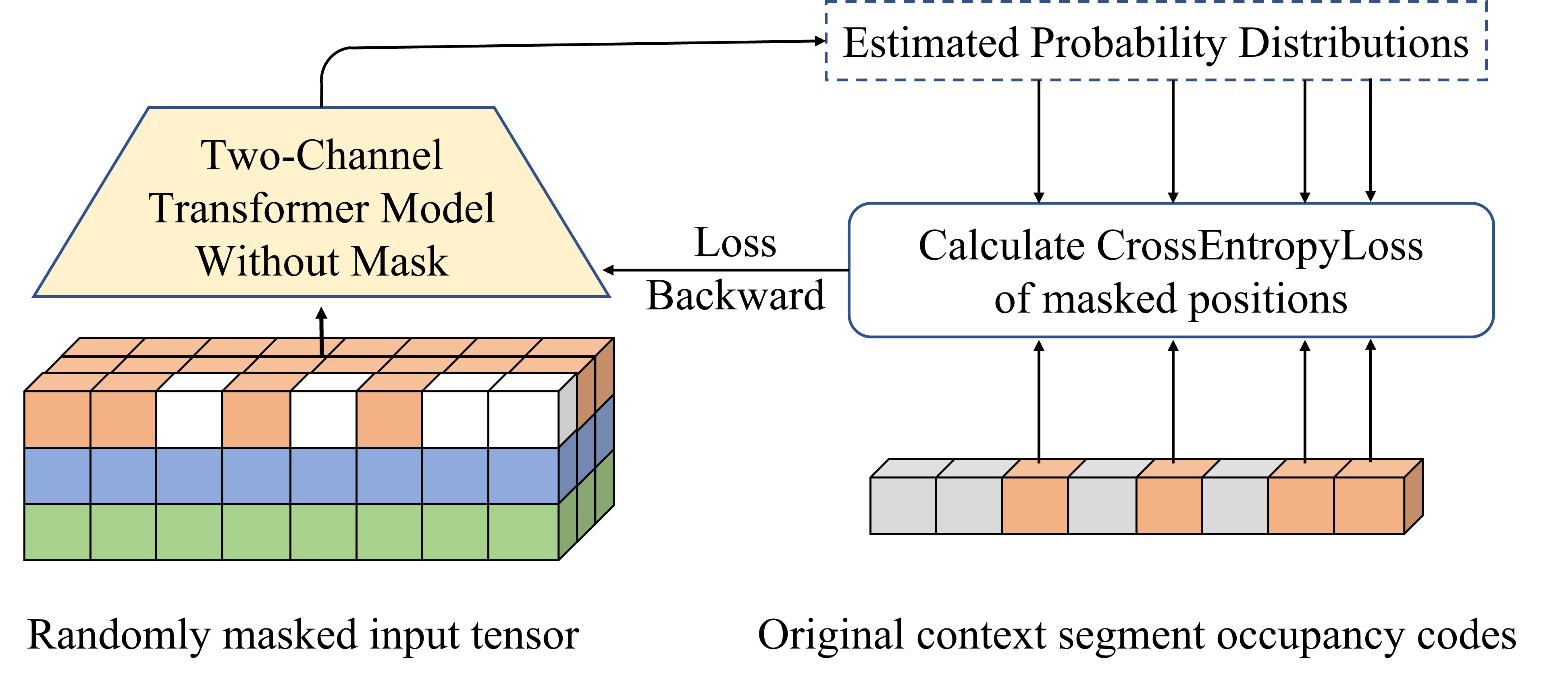}
    \caption{Overview of our pre-train strategy. We use same model without mask to process input, and only backward the loss at masked positions.}
    \label{pretrain}
\end{figure}
The essential task of our transformer is to use context $\bm{x}^{<i},\bm{y}^{\le i},\bm{x}^{i}_{w_j^{<k}}$ to predict current group $\bm{x}^{i}_{w_j^k}$. Therefore, it is important to fully exploit the predict ability of our transformer-based model. Inspired by BERT \cite{BERT}, we design a pre-train method to help our model develop the ability of predicting current node with context information (Fig. \ref{pretrain}). Sepcifically, we randomly mask the input occupancy code with $50$\% probability, and use our model (without multi-group mask) to predict the masked positions with unmasked occupancy codes and full information of ancestors. This operation simulates the process of actual coding. As for gradient backward, we only backward the loss between predicted and ground truth occupancy codes of masked positions. %We employ our pre-train method on object point cloud compression, and pre-trained for 2 epochs. We test our pre-train method on object dataset, and 
%We achieve considerable bpp improvement (See Tab. \ref{OctAttentionCompare}).

\section{Experiment}
\begin{figure*}[h]
\centering
\includegraphics[width=\textwidth]{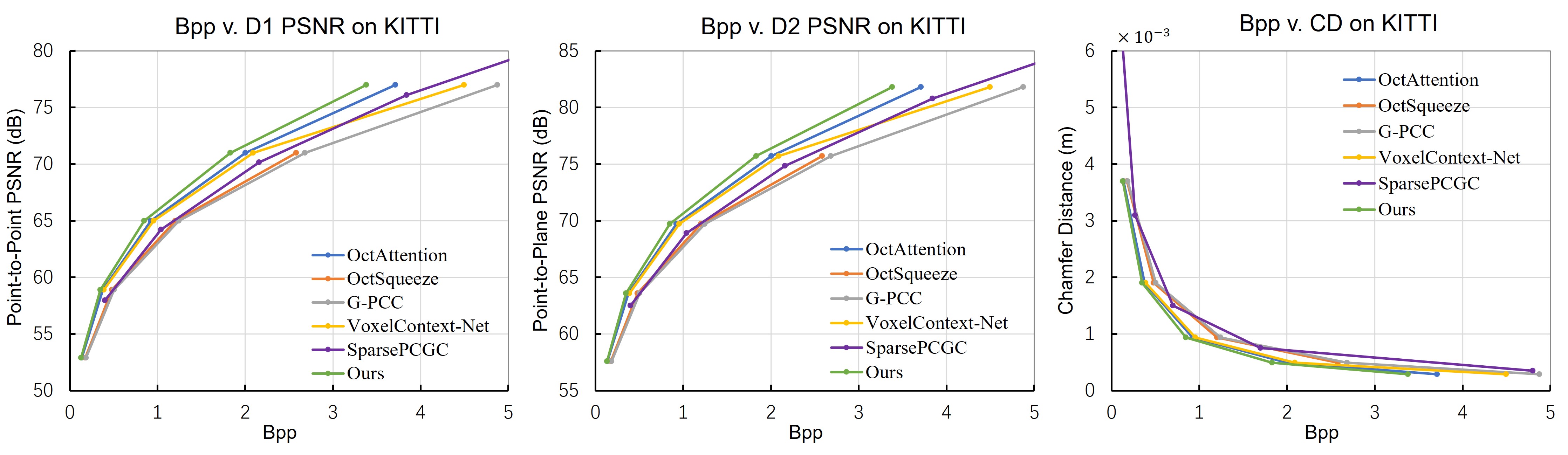}
\caption{Results of different methods on SemanticKITTI dataset.}
\label{Lidarresults}
\end{figure*}
\subsection{Datasets}
We train and test our method on LiDAR and object dataset for comprehensive evaluation.

For LiDAR dataset, we adopt SemanticKITTI \cite{KITTIdataset}, which is a large-scale outdoor-scene dataset for sparse point cloud semantic segmentation. It consists of 22 sequences, and contains a total of 43552 scans with 4549 million points.

For Object dataset, we adopt 8i Voxelized Full Bodies (MPEG 8i) \cite{MPEGdataset} and Microsoft Voxelized Upper Bodies (MVUB) \cite{MVUBdataset}. MPEG 8i contains sequences of smooth surface and complete human shape point clouds with 10 and 12bit precision. MVUB contains sequences of five half-body human shape dynamic voxelized point cloud with 9 and 10-bit precision. Due to their fixed precision, MPEG 8i and MVUB are commonly used for lossless compression task to fairly compare the compression rate between different methods.

To ensure fair comparison, we use the same train-test split as previous works \cite{fu2022octattention, que2021voxelcontext}, and perform lossy compression task on LiDAR dataset, lossless compression task on object datasets.

\subsection{Implementation Details}

\subsubsection{Experimental Setup}
All experiments are performed on a machine with NVIDIA A100-PCIE-40GB GPU. We also test our method and OctAttention on the machine with one NVIDIA GeForce RTX 3090 for fair comparison with results reported in SparsePCGC \cite{wang2021sparse}. We use Adam optimizer with $lr=10^{-3}$. It takes 2-4 days to train our model to 20 epochs. We tried different context segment length $n$ and group number $g$ for ablation study. To balance compression bitrate and decoding time, we set $n=1024,g=8$ for LiDAR model and $n=2048,g=8$ for object model as default settings.

\subsubsection{Baseline Methods}
In LiDAR point cloud lossy compression task, we set the maximum level of octree from 8 to 12 to achieve different R-D trade-off. We compare our work with octree-based compression method OctAttention \cite{fu2022octattention}, VoxelContext-Net \cite{que2021voxelcontext} OctSqueeze \cite{octsq}, competitive voxel-based learned method SparsePCGC \cite{wang2021sparse}, and traditional hand-crafted G-PCC from MPEG standard in the stable version (TMC13 v14.0) \cite{gpcc}. %Other methods focusing on LiDAR compression ~\cite{riddle, wang2022point, he2022density} are not feasible in comparison, for they focus on lower PSNR range of compression. 
In object point cloud compression, we set the side length of the leaf cube to 1 to achieve lossless compression. We compare our method with G-PCC, OctAttention, SparsePCGC \cite{wang2021sparse}, VoxelDNN \cite{nguyen2021learning}, MSVoxelDNN \cite{nguyen2021multiscale}. We apply lossless compression on above methods for fair comparison on bpp and encoding/decoding time. 

\subsubsection{Training and Testing Strategy}
Following previous octree-based works, we train 2 separate model on LiDAR and object datasets. For LiDAR task, we construct the training point could to 12-level octree so that our model can learn the distribution at all level in one training procedure. We truncate the octree to 8-12 levels to test our model at different bitrates. For object point clouds, we construct them losslessly to octree for training and testing.

\subsubsection{Evaluation Metrics}
For fair comparison, we adopt same evaluation metrics with previous works \cite{fu2022octattention, que2021voxelcontext}. Specifically, for LiDAR, we use point-to-point (D1 PSNR), point-to-plane (D2 PSNR) and chamfer distance (CD) to measure reconstruction quality in lossy compression. We compare bpp, encoding time and decoding time to evaluate compression performance. The results of compared works are performed on machines with the same GPU configuration. All bpp data are obtained by averaging over sequence. 

\subsection{Experimental Results}
\subsubsection{Lossy Compression Performance}
The bpp-distortion curves of LiDAR point cloud compression is shown in Fig. \ref{Lidarresults}. Our model outperforms other methods at all bitrates consistently. Specifically, we save $31$\% bitrate over G-PCC averagely over five distortion levels. We also achieve $8.9$\% and $4.4$\% relative reduction at high and low bitrate respectively versus SOTA method OctAttention. Experimental results verify the effectiveness of our multi-group context model over previous methods. 

\begin{table*}[htbp]
\centering
\begin{tabular}{@{}lllllll@{}}
\toprule
\multirow{2}{*}{Point Cloud} & Traditional & \multicolumn{3}{c}{Voxel-based} & \multicolumn{2}{c}{Octree-based}            \\ \cmidrule(lr){2-2} \cmidrule(lr){3-5} \cmidrule(lr){6-7} 
                             & G-PCC       & SparsePCGC  & VoxelDNN & MSVoxelDNN & OctAttention & Ours \\ \midrule
 loot\_vox10 (bpp) & 0.95 & 0.63 & 0.58 & 0.73 & 0.62 & 0.55 \\
 redandblack\_vox10 (bpp) & 1.09 & 0.72 & 0.66 & 0.87 & 0.73 & 0.66 \\
 boxer\_viewdep\_vox10 (bpp) & 0.94 & 0.60 & 0.55 & 0.70 & 0.59 & 0.51 \\
 Thaidancer\_viewdep\_vox10 (bpp) & 0.99 & 0.67 & 0.68 & 0.85 & 0.65 & 0.58 \\  \midrule
 Average bpp & 0.99 & 0.66 & 0.62 & 0.79 & 0.65 & 0.58 \\
 Average Gain over G-PCC &  - & 33.8\% & 37.6\% & 20.5\% & 34.6\% & 41.9\% \\ \midrule
 Average Encoding Time (s) & 4.0 & 9.5 & 885 & 54 & 0.80 & 1.92 \\
 Average Decoding Time (s) & 1.0 & 9.1 & 640 & 58 & 948 & 19.5 \\ \bottomrule
~\\
\end{tabular}

\caption{Bpp and coding time results on MPEG 8i dataset compared with G-PCC, voxel-based and octree-based methods.} %, all data is performed on same hardware configuration.}
\label{bppandtimetable}
\end{table*}

\subsubsection{Lossless Compression Performance}
The lossless compression results on object datasets are shown in Table \ref{Objresults}, our method saves $42.4$\% and $33.6$\% bpp averagely on MPEG 8i and MVUB over traditional method G-PCC. We also compare our method w/ and w/o pre-train. Tab.~\ref{OctAttentionCompare} shows that our method w/ pre-train outperforms OctAttention by $6.3$\% and $1.3$\% on two datasets, and gets considerable gains over models w/o pre-train.

\begin{table}[htbp]
\centering
\begin{tabular}{@{}lllll@{}}
\toprule
\multirow{2}{*}{Point Cloud} & \multicolumn{3}{c}{bpp}     & \multirow{2}{*}{Gain*} \\ \cmidrule(lr){2-4}
                             & G-PCC & OctAttention & Ours & \\  \midrule
\multicolumn{5}{@{}l@{}}{\textit{MPEG 8i}} \\
Loot10          & 0.95 & 0.62 & 0.55 & 42.1\% \\
Redandblack10   & 1.09 & 0.73 & 0.66 & 39.4\% \\
Boxer9          & 0.96 & 0.60 & 0.53 & \multirow{2}{*}{45.3\%} \\
Boxer10         & 0.94 & 0.59 & 0.51 &                   \\
Thaidancer9     & 0.99 & 0.64 & 0.59 & \multirow{2}{*}{40.9\%} \\
Thaidancer10    & 0.99 & 0.65 & 0.58 &                   \\
Average         & 0.99 & 0.64 & 0.57 & 42.4\% \\ \midrule
\multicolumn{5}{@{}l@{}}{\textit{MVUB}}    \\
Phil9           & 1.23 & 0.83 & 0.79 & 35.8\% \\
Phil10          & 1.07 & 0.79 & 0.76 & 29.0\% \\
Ricardo9        & 1.04 & 0.79 & 0.76 & 34.6\% \\
Ricardo10       & 1.07 & 0.72 & 0.69 & 35.5\% \\
Average         & 1.10 & 0.76 & 0.73 & 33.6\% \\ \bottomrule
~\\
\end{tabular}
\caption{Lossless compression results of different methods on MPEG 8i and MVUB datasets. * The bpp saving over G-PCC.}
\label{Objresults}
\end{table}

\begin{table}[htbp]
\centering
\begin{tabular}{@{}lllll@{}}
\toprule
\multirow{2}{*}{Point Cloud} & \multicolumn{2}{l}{Ours (w/o pre-train)} & \multicolumn{2}{l}{Ours (w/ pre-train)} \\ \cmidrule(lr){2-3} \cmidrule(lr){4-5} 
                             & bpp               & Gain*               & bpp               & Gain*              \\ \midrule
MPEG 8i                      & 0.61               & 4.7\%                   & 0.60               & 6.3\%              \\
MVUB                         & 0.77               & -1.3\%                   & 0.75               & 1.3\%              \\ \bottomrule
~\\
\end{tabular}
\caption{Ablation study of lossless compression on MPEG 8i and MVUB dataset. * The bpp saving over OctAttention.}
\label{OctAttentionCompare}
\end{table}

\subsubsection{Encoding and Decoding Efficiency}
We adopt MPEG 8i dataset to compare encoding and decoding efficiency. And we compare our results against G-PCC, voxel-based method SparsePCGC \cite{wang2021sparse}, VoxelDNN\cite{nguyen2021learning}, and  octree-based method OctAttention \cite{fu2022octattention}. As shown in Tab.~\ref{bppandtimetable}, our method outperforms other deep learning methods in compression performance, and saves 98\% of decoding time compared with OctAttention. The encoding time of our method is slower than Octattention due to the dual transformer structure. Moreover, a scatter plot of bpp and decoding time is shown in Fig. \ref{bpptime}. It can be seen that our method achieves SOTA compression performance with very fast decoding time.

\subsection{Ablation Study and Analysis}
\subsubsection{Dual Branch Transformer}
To explore the effectiveness of our dual branch transformer, we separate the two branches into independent models and test them separately. We adopt MPEG 8i dataset and test average bpp, encoding time, total decoding time and neural network inference time to evaluate the performance. We set context segment length $n=2048$ and group count $g=8$ for group-parallel branch, we train $20$ epochs for each model. Tab.~\ref{twochannels} demonstrates that the bitrate results obtained by the separate models are worse than the dual model. Meanwhile, both branches are marginally accelerated in decoding, as the size of neural network is reduced. The level-parallel branch can decode the context segment in parallel, so that the running time of neural network is negligible. Experimental results verify that the dual branch approach is important to compression performance while brings only marginal decoding time overhead.

\subsubsection{Effect of Context Segment Length}
We set the context segment length $n\in \{128,256,512,1024,2048,4096\}$ to study the effect of $n$. For each model we set group count $g=8$ and train $20$ epochs. And all other parameters remain the same. Tab.~\ref{contextlength} shows we obtain 12.1\% reduction on bitrate by enlarging $n$ from $128$ to 2048, this is because more context information can be used in a larger segment. Moreover, because expanding the context segment size $n$ can reduce the number of forward pass and enable our device to process data continuously, we reduce the encoding time by $7.5$ times, and decoding time by $8.4 $ times. And we also reduce neural network inference time, which decreases by $12$ times while decoding. Therefore, enlarging context segment will be more effective and also more efficient in our method. We set $n=2048$ for best performance under the constrain of GPU memory. We do not implement multi-threading in our paper, the impact of $n$ on efficiency only holds under this condition, and the speedup in Tab.~\ref{bppandtimetable} is brought by the reduction of serial decoding complexity.

\begin{table}[htbp]
\centering
\begin{tabular}{@{}lllll@{}}
\toprule
\multirow{2}{*}{Branch} & \multirow{2}{*}{\begin{tabular}[l]{@{}l@{}}Avg. bpp on\\ MPEG 8i\end{tabular}} & \multirow{2}{*}{\begin{tabular}[l]{@{}l@{}}Encoding\\ time (s)\end{tabular}} & \multicolumn{2}{l}{Decoding time (s)} \\ \cmidrule(lr){4-5} 
 &  &  & Total  & Network \\ \midrule
Level (L) & 0.75 & 0.79 & 7.2 & 0.8 \\
Group (G) & 0.64 & 0.82 & 9.4 & 4.0 \\
Dual (L+G) & 0.58 & 1.43 & 11.7 & 6.9 \\ \bottomrule
~\\
\end{tabular}%

\caption{Independent performance of two branches on object dataset MEPG 8i.}
\label{twochannels}
\end{table}

\begin{table}[htbp]
\centering
\begin{tabular}{@{}lllll@{}}
\toprule
\multirow{2}{*}{$n$} & \multirow{2}{*}{\begin{tabular}[l]{@{}l@{}}Avg. bpp on\\ MPEG 8i\end{tabular}} & \multirow{2}{*}{\begin{tabular}[l]{@{}l@{}}Encoding\\ time (s)\end{tabular}} & \multicolumn{2}{l}{Decoding time (s)} \\ \cmidrule(lr){4-5} 
 &  &  & Total  & Network \\ \midrule
128 & 0.66 & 10.8 & 98.8 & 84.3 \\
256 & 0.66 & 5.54 & 55.0 & 44.1 \\
512 & 0.61 & 2.68 & 29.9 & 22.5 \\
1024 & 0.59 & 1.46 & 16.4 & 11.0 \\
2048 & 0.58 & 1.43 & 11.7 & 6.9 \\
4096* & - & - & - & - \\\bottomrule
~\\
\end{tabular}%

\caption{Performance of bpp and coding time on object datatset MPEG 8i when using various context segment length. * We can not provide result with $n=4096$, as it exceeds $40$G GPU memory while training.}
\label{contextlength}
\end{table}

\begin{table}[htbp]
\centering
\begin{tabular}{@{}lllll@{}}
\toprule
\multirow{2}{*}{$g$} & \multirow{2}{*}{\begin{tabular}[l]{@{}l@{}}Avg. bpp on\\ MPEG 8i\end{tabular}} & \multirow{2}{*}{\begin{tabular}[l]{@{}l@{}}Encoding\\ time (s)\end{tabular}} & \multicolumn{2}{l}{Decoding time (s)} \\ \cmidrule(lr){4-5} 
 &  &  & Total  & Network \\ \midrule
2 & 0.66 & 1.38 & 7.9 & 2.6 \\
4 & 0.61 & 1.30 & 8.8 & 3.6 \\
8 & 0.58 & 1.43 & 11.7 & 6.9 \\
16 & 0.57 & 1.26 & 19.2 & 13.6 \\
32 & 0.57 & 1.37 & 33.3 & 26.6 \\ \bottomrule
~\\
\end{tabular}%

\caption{Performance of bpp and coding time on object datatset MPEG 8i when using various group count.}
\label{groupcount}
\end{table}

\subsubsection{Effect of Group Count}
We also study the effect of group count $g$ on the performance of our coding strategy. We set group count $g=\{2,4,6,8,16,32\}$, $n=2048$ and train the model for $20$ epochs. Tab.~\ref{groupcount} demonstrates that we save $13.6$\% bitrate by increasing group count from $2$ to $32$. This is because as the group count increases, more occupancy codes of sibling nodes can be used as guiding information by latter groups. The encoding time remains similar, as the encoding procedure is always in parallel. Also, because the decoding procedure is sequentially conducted by group, the decoding time shows a multiple upward trend with the increase of group count. It is also noteworthy that when $g\ge 8$, the gain in bpp becomes only marginal while the increment in decoding time becomes significant. We set $g=8$ as it is a sweet-spot with reasonably good compression performance and fast decoding time. 

\begin{figure}[htbp]
    \centering
    \includegraphics[width=\columnwidth]{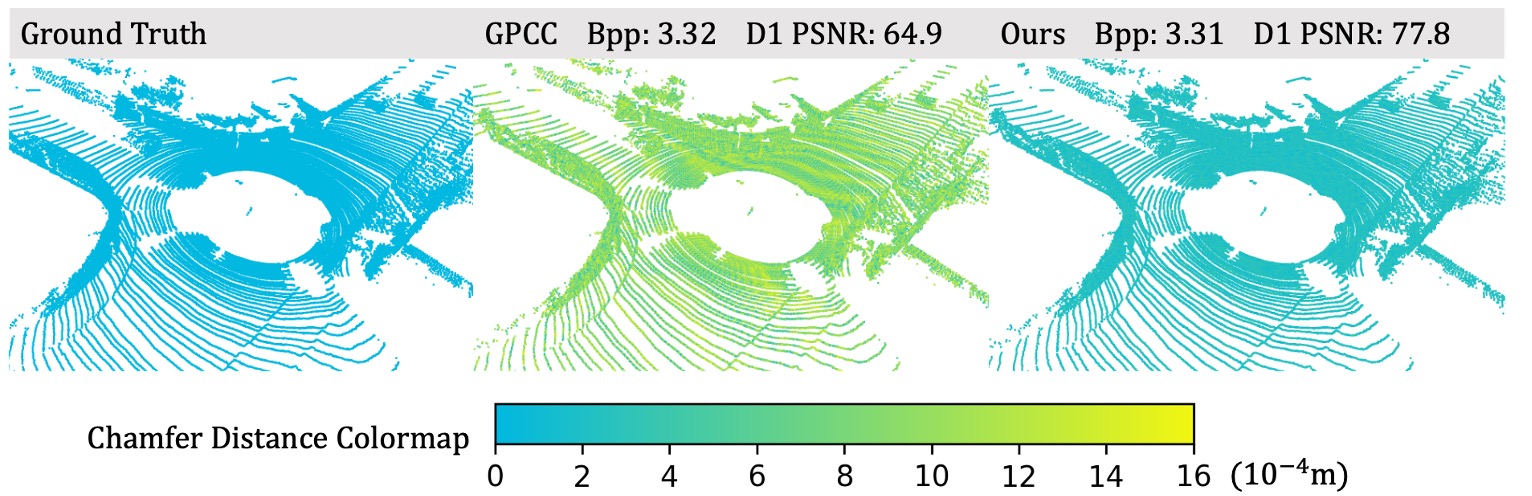}
    \caption{Chamfer distance between normalized ground truth and reconstructed point clouds on SemanticKITTI dataset.}
    \label{distortion}
\end{figure}

\subsection{Qualitative Results}
Fig.~\ref{distortion} shows the visualized  distortion of our method and G-PCC at similar compression rate. Our method outperforms G-PCC by more than 10 dB on D1 PSNR.

\section{Conclusion}
In this work, We propose an efficient large-scale context entropy model for point cloud geometry compression. To be specific, we propose a multi-group coding strategy to encode and decode the octree efficiently, based on which we propose a dual transformer architecture. We also design a random masking pre-train strategy. Results show that our model achieves SOTA compression performance, and reduce decoding time by up to 98\% compared to previous octree-based works (OctAttention~\cite{fu2022octattention}), which makes the practical deployment of octree-based DPCC possible. 

%\section{Discussion}
%The compression rate improvement over previous method on MVUB dataset is marginal compared with those on MPEG 8i and SemanticKITTI. It is unclear why the compression rate improvement on the three considered datasets is different. We leave this for future investigation.  %A possible solution is to investigate the optimal information bottleneck for lossy point cloud compression instead of fully relying on the quantization step in octree construction. We leave this for future work.

%\section{Discussion}
\newpage

{\small
\bibliographystyle{ieee_fullname}
\bibliography{egbib}
}

\newpage



\section*{Supplementary material}

# Appendix to ECM-OPCC: Efficient Context Model for Octree-based Point Cloud Compression

## 1. Proof of Bitrate Lowerbound in Sec. 3.2

In this section, we formalize and proof the theoretical claim on the bitrate lowerbound that is informally given in Tab. 2 of Sec 3.2. For clarity, in this section, we omit $y^i$ from conditioning. As $y^i$ is fully deterministic given $x^{i-1}$, this omission does not impact the lowerbound of bitrate.

**Theorem 1.** *For fully autoregressive model, the lowerbound of bitrate $R(X)$ to encode $X$ as symbols $\{x_1^1, ..., x_{N^1}^1, ..., x_{N^l}^l\}$ with entropy model $P_\theta(x_1^1), ..., P_\theta(x_{N^1}^1 | x_{<N^1}^1), ..., P_\theta(x_{N^l}^l | x^{<l}, x_{<N^l}^l)$ is $\lceil \mathbb{E}[-\log P(X)] \rceil$, where $\lceil . \rceil$ is the upper rounding operator. This bound is achievable given perfect parameter estimation and the entropy model family covers original distribution.*

*Proof.* From optimal source coding lowerbound [1], we have $\mathbb{E}_P[-\log P_\theta(X)] \leq R(X)$, and $P_\theta(X) = P_\theta(x_1^1) \prod P_\theta(x_j^i | x^{<i}, x_{<j}^i)$. Then we have

$$\mathbb{E}_P[-\log P_\theta(X)]$$
$$= \mathbb{E}_P[-\log P(X)] + D_{KL}[P(X)||P_\theta(X)]$$
$$\geq \mathbb{E}_P[-\log P(X)] \quad (9)$$

, and the equality holds only when $D_{KL}[P(X)||P_\theta(X)] = 0$. In other words, when $P(X) = P_\theta(X)$, this lowerbound is achievable. □

**Theorem 2.** *For the proposed window-wise parallel autoregressive model, the lowerbound of bitrate to encode $X$ as symbols $\{x_{w_1}^1, ..., x_{w_{h_1}}^1, ..., x_{w_{h_l}}^l\}$ with entropy model $P_\theta(x_{w_1}^1), ..., P_\theta(x_{w_{h_1}}^1), ..., P_\theta(x_{w_{h_l}}^l | x^{<l}, x_{w_{<h_l}}^l)$ is $\lceil \mathbb{E}[-\log P(X)] + \sum_{ij}\mathbb{H}(x_{w_j}^i | x^{<i}) - \mathbb{H}(X) \rceil$, where $\lceil . \rceil$ is the upper rounding operator and $\mathbb{H}$ is the entropy. This bound is achievable given perfect parameter estimation, the entropy model family covers original distribution, the original distribution factorizes as window size and each group contains only 1 node.*

*Proof.* As we have

$$P_\theta(X) = P_\theta(x^1) \prod P_\theta(x^i | x^{<i}) \quad (10)$$
$$P_\theta(x^i | x^{<i}) = P_\theta(x_{w_1}^i | x^{<i}) \prod P_\theta(x_{w_j}^i | x^{<i}, x_{w_{<j}}^i) \quad (11)$$

From [5], we have:

$$D_{KL}[P(x^i | x^{<i})||P_\theta(x^i | x^{<i})]$$
$$\geq \sum_j \mathbb{H}(x_{w_j}^i | x^{<i}) - \mathbb{H}(x^i | x^{<i}) \quad (12)$$

Then from the clain rule of entropy and relative entropy [1], we have:

$$D_{KL}[P(X)||P_\theta(X)]$$
$$= \sum_i D_{KL}[P(x^i | x^{<i})||P_\theta(x^i | x^{<i})]$$
$$\geq \sum_i (\sum_j \mathbb{H}(x_{w_j}^i | x^{<i}) - \mathbb{H}(x^i | x^{<i}))$$
$$= \sum_i \sum_j \mathbb{H}(x_{w_j}^i | x^{<i}) - \mathbb{H}(X) \quad (13)$$

Then, taking Eq. 13 into Eq. 9 and following the proof of Thm. 1 complete the proof. □

## 2. More Related Works

In general, there are four types of DPCC frameworks: octree-based, voxel-based, projection-based and point-based approach. Octree-based compression is introduced in the main text. Voxel-based approaches [12] [13] quantize point cloud to voxel grids and use 3D auto-encoders to compress them. VoxelDNN [9, 10, 11] introduces hybrid compression by first building octree to store structural information of point cloud and then compressing local voxelized point cloud by 3D convolution. Another line of research [18, 17] also uses hybrid scheme but downsamples point cloud by learned-transform first and then utilizes tree-based approach to compress point cloud. Without autoregressive context modeling, these methods fail to generalize on sparse point cloud [3]. For sparse point cloud, projection

1

| Distortion | | | G-PCC | OctSqueeze | VoxelContex-net | OctAttention | Ours | Gain* |
|---|---|---|---|---|---|---|---|---|
| D1 PSNR | D2 PSNR | CD | | | | | | |
| 52.9 | 57.6 | 0.003700 | 0.185 | 0.177 | - | 0.137 | 0.131 | 29.19% |
| 58.9 | 63.6 | 0.001900 | 0.505 | 0.480 | 0.390 | 0.376 | 0.348 | 31.12% |
| 65.0 | 69.7 | 0.000937 | 1.245 | 1.202 | 0.956 | 0.924 | 0.847 | 31.95% |
| 71.0 | 75.7 | 0.000492 | 2.684 | 2.582 | 2.089 | 2.004 | 1.831 | 31.78% |
| 77.0 | 81.8 | 0.000290 | 4.874 | - | 4.497 | 3.711 | 3.380 | 30.65% |

Table 9. Detail results for LiDAR point cloud compression. *The bpp saving over G-PCC

based approaches [16, 15, 2, 20] first project point cloud to image then use image compression methods to compress then. This kind of method handles large sparse point cloud well and evolves as image compression methods improve. The above mentioned point cloud compression methods can compress point cloud in lossy and lossless way but do not preserve density in different regions of a point cloud. To tackle this problem, point-based methods [19, 4, 6, 7] utilize network designed for raw point cloud segmentation as transform and encode transformed points and features by arithmetic coding.

## 3. More Implementation Details

### 3.1. Datasets

To make fair comparison, we use the same training sequences as previous works [3, 14]. For object point clouds, we use Soldier10 and Longdress10 sequences in MPEG 8i, Andrew10, David10, Sarah10 sequences in MVUB for training. We select Loot10, Redandblack10, Boxer9, Boxer10, Thaidancer9, Thaidancer10 in MPEG 8i, Phil9, Phil10, Ricardo9, Ricardo10 in MVUB for testing. For LiDAR datasets, we use sequences 00 to 10 (23201 scans in total) for training, and 11 to 21 (20351 scans in total) for testing. All testing sequences were not used in training. Specifically, Boxer9, Boxer10, Thaidancer9, Thaidancer10 are sequences containing only one frame, we repeat the test 20 times and take the average coding time as results to avoid fluctuation.

### 3.2. Experimental Setup

We intended to perform previous methods G-PCC [8], VoxelDNN [9], MSVoxelDNN [11], SparsePCGC [17] and OctAttention [3] on our machine to make fair comparison on coding performance and efficiency. However, some of the works above do not have open-source code or have not provide standard trained models. Therefore, we use results from SparsePCGC [17], which contains bpp, encoding time and decoing time results of G-PCC, VoxelDNN, MSVoxelDNN and SparsePCGC on loot_vox10, redandblack_vox10, boxer viewdep_vox10 and Thaidancer_viewdep_vox10.

| CPU Specifications (s) | 8157 (Ours) | 6226R |
|---|---|---|
| Total Cores | 24 | 16 |
| Processor Base Frequency | 2.30 GHz | 2.90 GHz |

Table 10. CPU specifications of Intel Xeon Platinum 8157 and Intel Xeon 6226R. Data comes from Intel official website.

| Point Cloud | Encoding time (s) | Decoding time (s) |
|---|---|---|
| longdress_vox10_1300 | 5.5 | 3.8 |
| redandblack_vox10_1550 | 5.0 | 3.5 |
| soldier_vox10_0690 | 7.5 | 5.0 |
| loot_vox10_1200 | 5.9 | 4.1 |
| Average | 6.0 (*4.0) | 4.1 (*1.0) |

Table 11. Coding time of G-PCC on four point cloud frames. *The average results reported in SparsePCGC [17].

According to [17], these testings are performed on a machine with Intel Xeon 6226R CPU and a Nvidia GeForce RTX 3090 GPU. Therefore, to make fair comparison, we test our model and OctAttention on a machine with Intel Xeon Platinum 8157 CPU and a Nvidia GeForce RTX 3090 GPU. The comparison between our method and OctAttention is completely reliable. We do the following illustration to show that the comparison between our results and results in SparsePCGC is promising. The only difference on hardware configuration bewteen two machines is CPU. The specifications of two CPUs is shown in Tab. 10. Data show that the computing ability of two CPUs is at same level. To make further confirmation, we test G-PCC (only use CPU to compute) coding time on our machine, the coding time is shown in Tab. 11. Results demonstrate that our machine do not take advantage over CPU computing speed.

We perform our ablation experiments on a machine with AMD EPYC 7742 CPU and NVIDIA A100-PCIE-40GB GPU. The results performed on this machine are only used in ablation analysis to avoid unfair comparison.

| Method | Parameter Count | FLOPs |
|---|---|---|
| OctAttention | 4.2 M | 7.77 G |
| ECM-OPCC | 8.0 M | 18.09 G |

Table 12. Model size (parameter count) and FLOPs (for single forward computation) comparison between our method and OctAttention, the length of input is 1024.

| Point Cloud | OctAttention | Ours (w/o pre-train) | Ours (w/ pre-train) |
|---|---|---|---|
| *MPEG 8i* | | | |
| Loot10 | 0.62 | 0.59 | 0.58 |
| Redandblack10 | 0.73 | 0.70 | 0.69 |
| Boxer9 | 0.60 | 0.57 | 0.56 |
| Boxer10 | 0.59 | 0.55 | 0.54 |
| Thaidancer9 | 0.64 | 0.64 | 0.62 |
| Thaidancer10 | 0.65 | 0.62 | 0.61 |
| Average | 0.64 | 0.61 | 0.60 |
| *MVUB* | | | |
| Phil9 | 0.83 | 0.82 | 0.80 |
| Phil10 | 0.79 | 0.80 | 0.78 |
| Ricardo9 | 0.72 | 0.72 | 0.70 |
| Ricardo10 | 0.72 | 0.73 | 0.72 |
| Average | 0.76 | 0.77 | 0.75 |

Table 13. Complete bpp results on w/o pre-train and w/ pre-train models compared against OctAttention

# 4. More Experimental Results

## 4.1. Model Size and Computation Cost

Tab. 12 shows the model size and computation cost (FLOPs) of our method and OctAttention. Our model costs more on a single forward computation due to the dual-channel structure, but the time we need to call the model to compress a point cloud is greatly reduced due to our coding strategy.

## 4.2. Results for lossless compression on LiDAR

We use bpp results from previous works since some of the compared methods do not have open-source code. The bpp results of previous works could be used directly, for bpp results of same method stay the same while conducting on different machines. The complete results are shown in Tab. 9 We draw the data in Tab. 9 as Fig. 7 in the main text.

## 4.3. Results for Comparison against OctAttention

We set context window length $n = 1024$, and trained our model for 8 epochs to specifically compare with OctAttention, which use above settings. We test w/ and w/o pre-training step to vetify the effectiveness of pre-train method. The complete bpp results are shown in Tab. 13.

| | Level-Parallel Branch | | | |
|---|---|---|---|---|
| Point Cloud | Avg. Bpp on MPEG 8i | Encoding time (s) | Decoding time (s) | |
| | | | Total | Network |
| Loot10 | 0.72 | 0.93 | 7.90 | 0.92 |
| Redandblack10 | 0.83 | 0.86 | 7.70 | 0.90 |
| Boxer9 | 0.71 | 0.34 | 3.00 | 0.37 |
| Boxer10 | 0.67 | 1.17 | 10.90 | 1.27 |
| Thaidancer | 0.81 | 0.32 | 2.97 | 0.37 |
| Thaidancer10 | 0.75 | 1.13 | 10.70 | 1.23 |
| Average | 0.75 | 0.79 | 7.20 | 0.84 |

Table 14. Level-parallel branch, context window length $n = 2048$, training for 20 epochs.

## 4.4. Results of Dual Branch Ablation Study

The complete results of separate branches and dual model is shown in Tab. 14 to 16. We set context window length $n = 2048$ for three models, group count $g = 8$ for group-parallel branch and dual model. We train each model in ablation study for 20 epochs.

| | Group-Parallel Branch | | | |
|---|---|---|---|---|
| Point Cloud | Avg. Bpp on MPEG 8i | Encoding time (s) | Decoding time (s) | |
| | | | Total | Network |
| Loot10 | 0.62 | 0.96 | 11.1 | 4.6 |
| Redandblack10 | 0.73 | 0.89 | 10.3 | 4.3 |
| Boxer9 | 0.60 | 0.35 | 3.8 | 1.7 |
| Boxer10 | 0.59 | 1.20 | 14.0 | 5.9 |
| Thaidancer | 0.65 | 0.33 | 3.7 | 1.6 |
| Thaidancer10 | 0.65 | 1.17 | 13.4 | 5.7 |
| Average | 0.64 | 0.82 | 9.4 | 4.0 |

Table 15. Group-parallel branch, context window length $n = 2048$, group count $g = 8$, training for 20 epochs.

| | Dual model | | | |
|---|---|---|---|---|
| Point Cloud | Avg. Bpp on MPEG 8i | Encoding time (s) | Decoding time (s) | |
| | | | Total | Network |
| Loot10 | 0.56 | 1.71 | 14.0 | 8.1 |
| Redandblack10 | 0.67 | 1.57 | 12.9 | 7.5 |
| Boxer9 | 0.54 | 0.60 | 4.6 | 2.9 |
| Boxer10 | 0.53 | 2.08 | 17.1 | 10.1 |
| Thaidancer | 0.60 | 0.58 | 4.6 | 2.8 |
| Thaidancer10 | 0.59 | 2.02 | 17.0 | 10.0 |
| Average | 0.58 | 1.43 | 11.7 | 6.9 |

Table 16. Dual model, context window length $n = 2048$, group count $g = 8$, training for 20 epochs.

## 4.5. Results of Window Length Ablation Study

We use dual model, set group count $g = 8$, and change context window length $n \in$

$\{128, 256, 512, 1024, 2048, 4096\}$. The complete results of each model is shown in Tab. 17 to 21.

| | Context Window Ablation Test $n = 128$ | | | |
|---|---|---|---|---|
| Point Cloud | Avg. Bpp on MPEG 8i | Encoding time (s) | Decoding time (s) | |
| | | | Total | Network |
| Loot10 | 0.64 | 12.9 | 117.4 | 100.3 |
| Redandblack10 | 0.75 | 11.8 | 106.0 | 90.6 |
| Boxer9 | 0.62 | 4.2 | 38.1 | 32.6 |
| Boxer10 | 0.60 | 16.1 | 148.3 | 126.3 |
| Thaidancer | 0.68 | 4.0 | 37.1 | 31.8 |
| Thaidancer10 | 0.67 | 15.8 | 146.1 | 124.2 |
| Average | 0.66 | 10.8 | 98.8 | 84.3 |

Table 17. Dual model, context window length $n = 128$, group count $g = 8$, training for 20 epochs.

| | Context Window Ablation Test $n = 256$ | | | |
|---|---|---|---|---|
| Point Cloud | Avg. Bpp on MPEG 8i | Encoding time (s) | Decoding time (s) | |
| | | | Total | Network |
| Loot10 | 0.63 | 6.54 | 64.5 | 52.1 |
| Redandblack10 | 0.75 | 6.04 | 60.1 | 48.2 |
| Boxer9 | 0.62 | 2.11 | 21.3 | 17.1 |
| Boxer10 | 0.59 | 8.30 | 82.6 | 66.1 |
| Thaidancer | 0.68 | 2.12 | 20.7 | 16.6 |
| Thaidancer10 | 0.66 | 8.10 | 80.5 | 64.7 |
| Average | 0.66 | 5.54 | 55.0 | 44.1 |

Table 18. Dual model, context window length $n = 256$, group count $g = 8$, training for 20 epochs.

| | Context Window Ablation Test $n = 512$ | | | |
|---|---|---|---|---|
| Point Cloud | Avg. Bpp on MPEG 8i | Encoding time (s) | Decoding time (s) | |
| | | | Total | Network |
| Loot10 | 0.59 | 3.19 | 35.5 | 26.6 |
| Redandblack10 | 0.70 | 2.91 | 32.8 | 24.5 |
| Boxer9 | 0.57 | 1.06 | 12.0 | 8.9 |
| Boxer10 | 0.55 | 3.97 | 43.6 | 33.2 |
| Thaidancer | 0.63 | 1.04 | 11.2 | 8.4 |
| Thaidancer10 | 0.62 | 3.93 | 44.0 | 33.1 |
| Average | 0.61 | 2.68 | 29.9 | 22.5 |

Table 19. Dual model, context window length $n = 512$, group count $g = 8$, training for 20 epochs.

### 4.6. Results of Group Count Ablation Study

We use dual model, set context window length $n = 2048$, and change group count $g \in \{2, 4, 8, 16, 32\}$. The complete results of each model is shown in Tab. 22 to 26.

| Context Window Ablation Test $n = 1024$ | | | | |
|---|---|---|---|---|
| Point Cloud | Avg. Bpp on MPEG 8i | Encoding time (s) | Decoding time (s) | |
| | | | Total | Network |
| Loot10 | 0.57 | 1.70 | 19.3 | 13.0 |
| Redandblack10 | 0.68 | 1.57 | 17.9 | 12.0 |
| Boxer9 | 0.55 | 0.58 | 6.2 | 4.3 |
| Boxer10 | 0.53 | 2.17 | 25.1 | 16.5 |
| Thaidancer | 0.61 | 0.58 | 6.2 | 4.2 |
| Thaidancer10 | 0.60 | 2.14 | 23.5 | 15.8 |
| Average | 0.59 | 1.46 | 16.4 | 11.0 |

Table 20. Dual model, context window length $n = 1024$, group count $g = 8$, training for 20 epochs.

| Context Window Ablation Test $n = 2048$ | | | | |
|---|---|---|---|---|
| Point Cloud | Avg. Bpp on MPEG 8i | Encoding time (s) | Decoding time (s) | |
| | | | Total | Network |
| Loot10 | 0.56 | 1.71 | 14.0 | 8.1 |
| Redandblack10 | 0.67 | 1.57 | 12.9 | 7.5 |
| Boxer9 | 0.54 | 0.60 | 4.6 | 2.9 |
| Boxer10 | 0.53 | 2.08 | 17.1 | 10.1 |
| Thaidancer | 0.60 | 0.58 | 4.6 | 2.8 |
| Thaidancer10 | 0.59 | 2.02 | 17.0 | 10.0 |
| Average | 0.58 | 1.43 | 11.7 | 6.9 |

Table 21. Dual model, context window length $n = 2048$, group count $g = 8$, training for 20 epochs.

| Group Count Ablation Test $g = 2$ | | | | |
|---|---|---|---|---|
| Point Cloud | Avg. Bpp on MPEG 8i | Encoding time (s) | Decoding time (s) | |
| | | | Total | Network |
| Loot10 | 0.63 | 1.67 | 9.3 | 3.0 |
| Redandblack10 | 0.74 | 1.54 | 8.6 | 2.8 |
| Boxer9 | 0.62 | 0.60 | 3.1 | 1.0 |
| Boxer10 | 0.59 | 2.10 | 11.8 | 3.8 |
| Thaidancer | 0.69 | 0.58 | 3.0 | 1.0 |
| Thaidancer10 | 0.66 | 1.78 | 11.7 | 3.7 |
| Average | 0.66 | 1.38 | 7.9 | 2.6 |

Table 22. Dual model, context window length $n = 2048$, group count $g = 2$, training for 20 epochs.

| Group Count Ablation Test $g = 4$ | | | | |
|---|---|---|---|---|
| Point Cloud | Avg. Bpp on MPEG 8i | Encoding time (s) | Decoding time (s) | |
| | | | Total | Network |
| Loot10 | 0.59 | 1.57 | 10.2 | 4.2 |
| Redandblack10 | 0.70 | 1.46 | 9.5 | 3.9 |
| Boxer9 | 0.57 | 0.57 | 3.5 | 1.5 |
| Boxer10 | 0.55 | 1.98 | 13.1 | 5.3 |
| Thaidancer | 0.64 | 0.55 | 3.4 | 1.5 |
| Thaidancer10 | 0.62 | 1.66 | 12.8 | 5.2 |
| Average | 0.61 | 1.30 | 8.8 | 3.6 |

Table 23. Dual model, context window length $n = 2048$, group count $g = 4$, training for 20 epochs.

| Group Count Ablation Test $g = 8$ | | | | |
|---|---|---|---|---|
| Point Cloud | Avg. Bpp on MPEG 8i | Encoding time (s) | Decoding time (s) | |
| | | | Total | Network |
| Loot10 | 0.56 | 1.71 | 14.0 | 8.1 |
| Redandblack10 | 0.67 | 1.57 | 12.9 | 7.5 |
| Boxer9 | 0.54 | 0.60 | 4.6 | 2.9 |
| Boxer10 | 0.53 | 2.08 | 17.1 | 10.1 |
| Thaidancer | 0.60 | 0.58 | 4.6 | 2.8 |
| Thaidancer10 | 0.59 | 2.02 | 17.0 | 10.0 |
| Average | 0.58 | 1.43 | 11.7 | 6.9 |

Table 24. Dual model, context window length $n = 2048$, group count $g = 8$, training for 20 epochs.

| Group Count Ablation Test $g = 16$ | | | | |
|---|---|---|---|---|
| Point Cloud | Avg. Bpp on MPEG 8i | Encoding time (s) | Decoding time (s) | |
| | | | Total | Network |
| Loot10 | 0.55 | 1.60 | 22.8 | 16.0 |
| Redandblack10 | 0.66 | 1.48 | 20.8 | 14.8 |
| Boxer9 | 0.53 | 0.58 | 7.8 | 5.6 |
| Boxer10 | 0.52 | 1.37 | 28.5 | 20.2 |
| Thaidancer | 0.59 | 0.56 | 7.7 | 5.5 |
| Thaidancer10 | 0.58 | 1.96 | 27.7 | 19.7 |
| Average | 0.57 | 1.26 | 19.2 | 13.6 |

Table 25. Dual model, context window length $n = 2048$, group count $g = 16$, training for 20 epochs.

| Group Count Ablation Test $g = 32$ | | | | |
|---|---|---|---|---|
| Point Cloud | Avg. Bpp on MPEG 8i | Encoding time (s) | Decoding time (s) | |
| | | | Total | Network |
| Loot10 | 0.55 | 1.60 | 39.1 | 31.2 |
| Redandblack10 | 0.66 | 1.49 | 36.2 | 28.9 |
| Boxer9 | 0.53 | 0.58 | 13.6 | 10.9 |
| Boxer10 | 0.51 | 2.02 | 49.1 | 39.3 |
| Thaidancer | 0.58 | 0.56 | 13.1 | 10.6 |
| Thaidancer10 | 0.57 | 1.97 | 48.6 | 38.6 |
| Average | 0.57 | 1.37 | 33.3 | 26.6 |

Table 26. Dual model, context window length $n = 2048$, group count $g = 32$, training for 20 epochs.



\end{document}